\documentclass[10pt,twocolumn,letterpaper]{article}

\usepackage{iccv}
\usepackage{times}
\usepackage{epsfig}
\usepackage{graphicx}
\usepackage{amsmath}
\usepackage{amssymb}

\usepackage{multirow}

\usepackage[pagebackref=true,breaklinks=true,letterpaper=true,colorlinks,bookmarks=false]{hyperref}

\iccvfinalcopy 

\def\iccvPaperID{5984} 

\ificcvfinal\fi
\begin{document}

\title{Extremely Weak Supervised Image-to-Image Translation for\\ Semantic Segmentation}
\newcommand{\aand}{\hspace{6mm}}
\author{Samarth Shukla\aand Luc Van Gool\aand Radu Timofte\\
Computer Vision Lab, ETH Zurich, Switzerland\\
{\tt\small \{samarth.shukla, vangool, radu.timofte\}@vision.ee.ethz.ch}
}


\maketitle

\begin{abstract}
Recent advances in generative models and adversarial training have led to a flourishing image-to-image (I2I) translation literature.
The current I2I translation approaches require training images from the two domains that are either all paired (supervised) or all unpaired (unsupervised).
In practice, obtaining paired training data in sufficient quantities is often very costly and cumbersome.
Therefore solutions that employ unpaired data, while less accurate, are largely preferred.
In this paper, we aim to bridge the gap between supervised and unsupervised I2I translation, with application to semantic image segmentation.
We build upon pix2pix and CycleGAN, state-of-the-art seminal I2I translation techniques.
We propose a method to select (very few) paired training samples and achieve significant improvements in both supervised and unsupervised I2I translation settings over random selection.
Further, we boost the performance by incorporating both (selected) paired and unpaired samples in the training process.
Our experiments show that an extremely weak supervised I2I translation solution using only one paired training sample can achieve a quantitative performance much better than the unsupervised CycleGAN model, and comparable to that of the supervised pix2pix model trained on thousands of pairs.
\end{abstract}



\section{Introduction}
\label{sec:introduction}

Image-to-Image (I2I) translation~\cite{Isola_2017_CVPR} deals with the problem of mapping an image from a source domain to a target domain. Under this broad definition, any task that requires per-pixel predictions on an image, such as semantic segmentation, image super-resolution and image inpainting may be viewed as an I2I translation problem. The training setting for most I2I translation approaches comprises either a set of paired samples~\cite{Isola_2017_CVPR, zhu2017toward, wang2018high} or a set of unpaired samples~\cite{Zhu_2017_ICCV,liu2017unsupervised}. We refer to these two training settings as supervised and unsupervised, respectively. While training with paired samples in the supervised setting typically results in better translation performance, it may be difficult or time consuming to assemble such a dataset with all paired samples. On the other hand, there may be situations wherein it is easier to obtain samples from both domains that are not necessarily paired. However, approaches to the I2I translation problem that train with only unpaired samples tend to be more loosely constrained, and the resulting translated images are often unable to retain the original content and structure.

In this work, we investigate whether the training in unpaired settings can be assisted by selecting a relatively small number of paired samples, using semantic segmentation as the application domain. Semantic segmentation is an important problem in computer vision which aims to infer dense pixel-level object class information in a natural image. This structured prediction problem may be viewed as an image-to-image (I2I) translation problem from the domain of natural images to the domain of semantic labels and can be learned in an end-to end manner.

Our objective is to obtain performance gains in unsupervised I2I translation tasks by selecting a small number of paired samples.
The most simple and straightforward technique of selecting a subset of paired samples would be to choose them at random.
However, such an approach may limit the diversity of chosen samples and their resemblance to the entire dataset.
A better selection strategy would be to ensure that the selected samples capture the greatest \textit{amount of information} of the dataset.
We hypothesize that this would lead to better learning and generalization -- choosing two diverse frames to annotate would provide a higher performance gain that annotating two frames that are semantically similar.

\begin{figure}[thp]
\centering
    \includegraphics[width=1.0\columnwidth,trim=10 150 20 0]{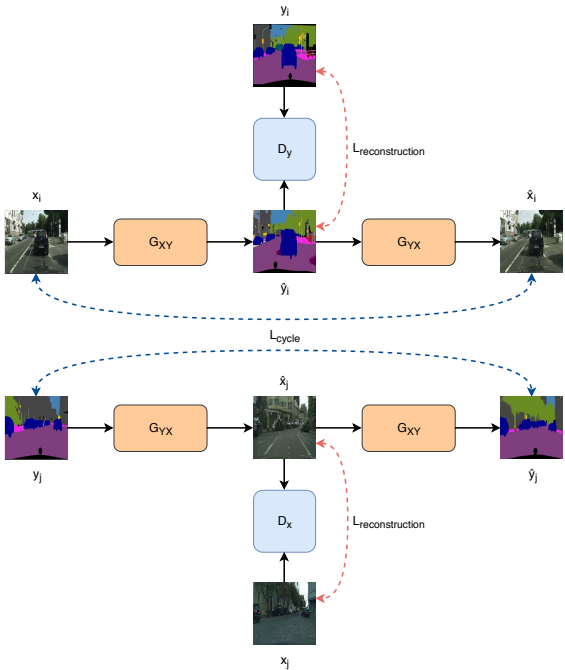}
    \caption{Architecture of our training process. Our model consists of two mapping functions, $G_{XY}: X\xrightarrow{}Y$ and $G_{YX}: Y\xrightarrow{}X$, and two discriminators $D_{X}$ and $D_{Y}$. When considering unpaired samples, we use the cycle-consistency loss between the original source image and the one obtained after first translating it to the target domain and then translating it back to the source domain. In case of paired samples, we also examine the reconstruction loss between the translated image and the corresponding target image in addition to the cycle consistency loss. (Figure best viewed in color and zoomed.)}
\label{fig:illustration}
\end{figure}

We aim to train an image-to-image translation model in settings where only a limited number of paired samples can be acquired.
Our main contributions are as follows:
\begin{itemize}
    \item We integrate widely used paired and unpaired image translation methods, namely pix2pix~\cite{Isola_2017_CVPR} and CycleGAN~\cite{Zhu_2017_ICCV} respectively, resulting in a framework where models can be trained using both paired and unpaired samples (see Fig.~\ref{fig:illustration}).
    \item We propose a method to select diverse samples from the dataset to be used in a paired setting, and experimentally show that it outperforms the approach where paired samples are randomly selected.
    \item We also demonstrate through quantitative evaluation that assisting the unpaired I2I translation problem with only a few paired images significantly improves performance.
    \item Additionally, for the unpaired setting, we use synthetically generated semantic labels from a different dataset and show that our method still achieves significant improvement in performance, a very practical feat.
\end{itemize}

\section{Related Work}
\label{sec:related-work}

\subsection{Generative Adversarial Networks}

Generative Adversarial Networks (GANs)~\cite{goodfellow2014generative} form a popular framework for generating realistic samples from high-dimensional complex data distributions.
It consists of two neural networks, referred to as a generator network, $G$, and a discriminator network $D$.
The two networks are trained to achieve mutually opposing goals in that $G$ attempts to generate samples from the real data distribution, whereas $D$ tries to distinguish between fake samples produced by $G$ and those from the real distribution.
This forces the generated samples to be indistinguishable from the real samples.
Recently, GANs have been applied successfully in several computer vision tasks such as conditional image generation~\cite{mirza2014conditional,reed2016generative}, image super-resolution~\cite{ledig2017photo}, object detection~\cite{li2017perceptual}, image inpainting~\cite{pathak2016context,yeh2017semantic} and domain adaptation~\cite{liu2016coupled,bousmalis2017unsupervised}.

\subsection{Supervised I2I Translation}
Prior to the advent of adversarial learning, learning of I2I translation tasks in neural networks was typically driven by per-pixel classification or regression losses~\cite{Long_2015_CVPR, Zhang_2016_ECCV}. These so-called \textit{unstructured} losses treat each pixel independently, thus ignoring any statistical dependencies among pixels.
Hand-crafted \textit{structured} losses such as conditional random fields~\cite{Chen_2015_ICLR} or the structural similarity metric (SSIM)~\cite{wang2004image} seek to account for such inter-pixel dependencies.

Adversarially-defined losses attempt to do away with cumbersome hand-crafting of loss functions in the same vein as end-to-end learning with neural networks does away with manually designed feature extraction.
Initial works that incorporated adversarial losses in I2I translation tasks~\cite{Pathak_2016_CVPR, Mathieu_2016_ICLR}, used them in combination with the aforementioned per-pixel regression losses.
Pix2pix~\cite{Isola_2017_CVPR}, a supervised I2I translation technique, employs a conditional GAN loss, in combination with a pixel-wise L1 loss.
A common drawback of all these methods is that they all require the training data to be paired between the source and the target domains.

\subsection{Unsupervised I2I Translation}
Obtaining per-pixel annotations is a challenging ask for a number of tasks and it would be highly desirable if comparable image translation performance can be achieved with no or minimal paired samples.
With this motivation, Zhu~\etal~\cite{Zhu_2017_ICCV} proposed the CycleGAN model, where the translation task is sought to be driven exclusively by the adversarial loss.
As such a loss seeks to match distributions rather than individual samples, paired training samples are not required in this approach.
However, the training of such a model is found to be unstable and the authors of~\cite{Zhu_2017_ICCV} resolve this by employing a cycle consistency regularization.
Specifically, they train two generators to map from either domain onto the other and require that the two mappings be bijections and themselves the inverses of each other.
This regularization is found to sufficiently constrain the optimization, and this model is widely considered to be the state of the art in unsupervised I2I translation.

\subsection{Semi-supervised I2I Translation}
Our work is most related to the work of Tripathy~\etal~\cite{tripathy2018learning}, where training of I2I translation models using a combination of paired and unpaired samples is proposed.
The main idea of their work is to define the cycle-consistency of a twice-translated image with respect to the original image not via a pixel-wise regression loss but instead through a separate discriminator conditioned on the once-translated image.
When paired samples are available, the discriminator is conditioned on the real image in the other domain instead of on the once-translated image.
They use a training strategy where all the paired samples are first used for learning for 50 epochs, and then the unpaired samples are used for next 150 epochs.
Our training strategy, on the other hand, utilizes both the paired and unpaired samples in each training epoch. Our experiments reveal that our simple approach outperforms in both labels$\xrightarrow{}$photos and photos$\xrightarrow{}$labels tasks, without using additional discriminators and with substantially fewer paired training samples.

\subsection{Feature Extraction}
ConvNets such as VGG~\cite{simonyan2014very} and ResNet~\cite{he2016deep}, trained on large scale Imagenet classification task have been used as feature extractors and successfully applied to various tasks, which include image super-resolution~\cite{ledig2017photo},~\cite{wang2018esrgan}, style transfer~\cite{Gatys_2016_CVPR, johnson2016perceptual} and semantic segmentation~\cite{chen2018deeplab}.
We follow these works in employing Imagenet pre-trained ResNet features as image representations which are then utilized for examining sample diversity, as explained in the following section.


\section{Proposed Method}

\subsection{Problem Formulation}
Given image samples from two domains $X$ and $Y$, our objective is to learn functions $G_{XY}$ and $G_{YX}$ which map images from one domain to the other, namely $\mathcal{X}\xrightarrow{}\mathcal{Y}$ and $\mathcal{Y}\xrightarrow{}\mathcal{X}$, respectively.
We assume that the majority of the training samples are unpaired and we are restricted to having only a limited number of paired samples.
In this setting, our approach seeks to address the following questions: a) how to choose which samples should be paired and b) how to integrate the training process to utilize both paired and unpaired samples.

\subsection{Selection of paired samples}
Given a specific limit to the number of samples to be selected and the corresponding pairs to be acquired, we seek to identify samples that capture the highest variance in the data.
In order to do so, we compute features of each image in the dataset by passing them through a pre-trained ResNet50~\cite{he2016deep} and consider the activations at the penultimate layer as a low-dimensional representation of the particular image.
Subsequently, we apply k-means clustering and assign a cluster label to each sample.
We choose the number of clusters to be equal to our fixed budget of number of paired samples that can be made available.
The representative sample of each cluster is then chosen to be the one with the least mean distance with all the other samples in that cluster -- the medoid sample.
In the following sections, we will refer to this strategy of selecting a fixed number of paired samples as \textit{k-medoids}.

\subsection{Learning Objective}
The learning objective of our model consists of the following loss components: an adversarial loss, a cycle-consistency loss, an identity loss and where applicable, the L1 reconstruction loss.

\subsubsection{Adversarial Loss }
The adversarial loss ensures that images generated by the generator look realistic.
We adopt the \textit{relativistic discriminator}~\cite{jolicoeur2018relativistic}, a recently proposed technique to improve training stability of GANs.
The relativistic discriminator predicts the probability of a real image being more realistic than a fake image instead of simply predicting the probability of the sample being real.
For the mapping for source domain, $X$, to target domain, $Y$, the discriminator and generator adversarial losses can respectively be written as:

\begin{equation}
    \mathcal{L}^{D}_{GAN}(G_{XY}, D^{Rel}_{Y}, X, Y) = \mathbb{E}_{y, \hat{y}}[logD^{Rel}_{Y}(y, \hat{y})]
\end{equation}
\begin{equation}
    \mathcal{L}^{G}_{GAN}(G_{XY}, D^{Rel}_{Y}, X, Y) = \mathbb{E}_{y, \hat{y}}[logD^{Rel}_{Y}(\hat{y}, y)]
\end{equation}
\\
where $y$ represents real samples,
$\hat{y}=G_{XY}(x)$ represents fake samples,
and $D^{Rel}_{Y}(y, \hat{y})=$ sigmoid$(C(y) - C(\hat{y}))$ and $C$ refers to the non-transformed output of the discriminator.

The above loss formulation for the generator uses real image samples as well, and thus encourages the generator to reduce the probability of the real data being real, along with increasing the probability of the fake data being real.

Similar formulation for adversarial losses
$\mathcal{L}^{D}_{GAN}(G_{YX}, D^{Rel}_{X}, X, Y)$ and
$\mathcal{L}^{G}_{GAN}(G_{YX}, D^{Rel}_{X}, X, Y)$ can be done for the inverse mapping from domain $Y$ to $X$.
We refer to the sum of these losses as $\mathcal{L}_{GAN}$.

\subsubsection{Cycle-consistency Loss~\cite{Zhu_2017_ICCV,Yi_2017_ICCV,kim2017learning}}
Intuitively, the mapping from domain
$X$ to $Y$, $G_{XY}: \mathcal{X}\xrightarrow{}\mathcal{Y} $
should be the inverse of mapping from domain
$Y$ to $X$, $G_{YX}: \mathcal{Y}\xrightarrow{}\mathcal{X} $, and vice-versa.
The cycle-consistency loss states that an image translated from source to target domain, and then, back to the source domain should lead to the original image. This loss forces the mapping between domain X and Y to be the inverse of mapping between domain Y and X by utilizing L1 difference between the reconstructed and original image. Formally, this loss is written as:
\begin{equation}
    \begin{split}
    \mathcal{L}_{cycle}(G_{XY}, G_{YX}) &=  \mathbb{E}_{x}||G_{YX}(G_{XY}(x)) - x||_{1}\\
    &+ \mathbb{E}_{y}||G_{XY}(G_{YX}(y)) - y||_{1}
    \end{split}
\end{equation}

Cycle-consistency loss tries to ensure that the fake image in the target domain when translated back to the source domain should result in a reasonably faithful reconstruction of the original image.
This may lead to the generators learning how to retrieve original source content and structure from fake target images.
However, the information required for this retrieval may be hidden in the fake images and there is no guarantee that this information will be explicitly encoded in the fake target images generated in the form of similar content and structure.

\subsubsection{Identity Loss}
The identity loss is written as:
\begin{equation}
    \begin{split}
    \mathcal{L}_{idt}(G_{XY}, G_{YX}) &=  \mathbb{E}_{x}||(G_{YX}(x) - x||_{1}\\
    &+ \mathbb{E}_{y}||G_{XY}(y) - y||_{1}
    \end{split}
\end{equation}

Intuitively, the identity loss is a regularization technique built on the idea that if an image from target domain is fed into a generator which maps from source to target domain, then the resulting mapping should be an identity~\cite{taigman2016unsupervised}.

\subsubsection{Reconstruction Loss for paired samples}
In cases where we have access to paired samples, we also consider a reconstruction loss along with the cycle consistency loss. This loss tries to minimize the L1 distance between the fake translated image generated in the target domain and its corresponding original image. A similar penalty is imposed for the inverse translation to the source domain.
This loss imposes a stricter constraint than the cycle-consistency loss and should lead to better preservation of image structure since deformations will be heavily penalized.
Mathematically, it is written as:
\begin{equation}
    \begin{split}
    \mathcal{L}_{L1}(G_{XY}, G_{YX}) &=  \mathbb{E}_{x}||(G_{XY}(x) - y||_{1}\\
    &+ \mathbb{E}_{y}||G_{YX}(y) - x||_{1}
    \end{split}
\end{equation}

\subsubsection{Total Loss}
The total objective comprises a weighted sum of these losses and can be written as:

\begin{equation}
    \mathcal{L}_{total} = \lambda_{1}\mathcal{L_{GAN}} +  \lambda_{2}\mathcal{L}_{cycle} +  \lambda_{3}\mathcal{L}_{idt} +  \lambda_{4}\mathcal{L}_{L1}
    \label{eq:total_loss}
\end{equation}

\subsection{Network Architectures}
In order to compare our results with previous works, we adopt the generator and discriminator architectures from CycleGAN~\cite{Zhu_2017_ICCV}.
We also train the pix2pix model~\cite{Isola_2017_CVPR} with these new architectures in order to establish a relevant baseline.

The generator network is based on~\cite{johnson2016perceptual,Zhu_2017_ICCV} and uses two stride-2 convolutions, nine residual blocks, and two fractionally strided convolutions with stride $\frac{1}{2}$, along with instance normalization. For the discriminator, we use PatchGANs~\cite{Isola_2017_CVPR} with $70\times70$ patch size.

\subsection{Training Details}

Image samples of size $256\times256$ are used for both the domains. The images are first re-scaled to size $286\times286$ and then a random crop of $256\times256$ is chosen. In addition, we randomly flip the images horizontally with a probability of 0.5. We use a batch size of 1 and employ the Adam optimizer~\cite{knigma2014adam}. The total number of epochs varies proportionally with the amount of samples used for training. All the generator and discriminator networks are trained with an initial learning rate of 0.0002, which is fixed for half the total epochs and linearly decays to zero in the subsequent epochs. We use a pool of size 50 of previously generated fake images to train the discriminators in order to stabilize training~\cite{shrivastava2017learning}. For the total loss~\eqref{eq:total_loss} we choose the weights $\lambda_1$, $\lambda_2$, $\lambda_3$, $\lambda_4$ to be 1, 10, 10 and 150, respectively.

\section{Experiments}
\label{sec:experiments}

\subsection{Cityscapes}
In this subsection, we compare the performance of our method with~\cite{tripathy2018learning}.
Our results show that our method performs better using less than 10\% of the training samples.

\subsubsection{Dataset}
The Cityscapes dataset~\cite{cordts2016cityscapes} consists of 3475 real-world driving images and corresponding pixel level semantic labels.
The dataset is split into 2975 images for training and 500 for validation.
Although the dataset consists of paired images, we consider the problem in an unpaired setting.

We train our models using 200 unpaired images selected using our k-medoids method, out of which a maximum of 40 images are chosen and used in a paired setting for our experiments. Using the trained models, we generate 500 fake images for photos$\xrightarrow{}$labels task.

\subsubsection{Evaluation protocol}
We adopt the approach presented in~\cite{long2015fully} to assess the quality of the synthesized images.
The standard metrics used are mean pixel accuracy, mean class accuracy and class IoU. For the photos$\xrightarrow{}$labels task, we first resize the generated image to the original resolution of Cityscapes ($1024\times2048$px). We then convert this image to class labels by utilizing the color map of each class provided in the cityscapes dataset and doing a nearest-neighbor search in color space, i.e., for each pixel in the image, we compute its distance with colors of each of the 19 class labels and assign it the class with which it has the lowest distance.
We then proceed to compute the metrics by comparing with ground truth labels.

\subsubsection{Selection of paired samples}
We first consider a fully-paired setting for the photos$\xrightarrow{}$labels task where we only have access to a limited number of samples. We vary the number of samples and conduct experiments on 1, 5, 10, 20 and 40 chosen randomly and using our k-medoids strategy. Table \ref{tab:results-kmedoids-vs-random-segmentation} shows performance of models trained using subsets of different number of samples selected using k-medoids and random criteria (average of 10 runs). We find that the models trained with samples chosen using the k-medoids criteria outperform the ones trained with samples selected randomly.

\begin{table}[ht]
\setlength{\tabcolsep}{4pt} 
\centering
\resizebox{\linewidth}{!}
{
\begin{tabular}[\linewidth]{|c|c||c c c||}
\hline
Samples & Selection & Pixel Acc. & Mean Acc. & Mean IoU\\
\hline\hline
\multirow{2}*{1} & random & 0.401 & 0.130 & 0.076 \\
& k-medoids & \textbf{0.567} & \textbf{0.160} & \textbf{0.114} \\
\hline
\multirow{2}*{5} & random & 0.633 & \textbf{0.190} & 0.139 \\
& k-medoids & \textbf{0.636} & 0.189 & \textbf{0.143} \\
\hline
\multirow{2}*{10} & random & 0.701 & 0.219 & 0.162 \\
& k-medoids & \textbf{0.720} & \textbf{0.229} & \textbf{0.174} \\
\hline
\multirow{2}*{20} & random & 0.720 & 0.237 & 0.189 \\
& k-medoids & \textbf{0.746} & \textbf{0.242} & \textbf{0.196} \\
\hline
\multirow{2}*{40} & random & 0.766 & 0.272 & 0.223 \\
& k-medoids & \textbf{0.781} & \textbf{0.278} & \textbf{0.236} \\
\hline\hline
2975 & - & 0.878 & 0.497 & 0.402\\
\hline
\end{tabular}
}
\caption{Average performance of models trained for photos$\xrightarrow{}$labels task using a subset of different sizes from the Cityscapes dataset (random vs k-medoids selection). For k-medoids selection, we take an average across 10 runs using the same samples, whereas for random selection, we take average over 10 runs using different random samples each time}
\label{tab:results-kmedoids-vs-random-segmentation}
\end{table}

Similar observations are made for the labels$\xrightarrow{}$photos task, the results for which can be found in the supplementary material.


\begin{table}[htp]
\setlength{\tabcolsep}{5pt}
\centering
\resizebox{\linewidth}{!}
{
\begin{tabular}{|c | c || c c c||}
\hline
Unpaired & Paired & Pixel Acc. & Mean Acc. & Mean IoU\\
\hline\hline
- & 200 & 0.830  & 0.366  & 0.299 \\
200 & - & 0.570  & 0.210 & 0.152 \\
\hline
\multirow{7}*{200 (\textbf{ours})}
& $1_{rand}$ & 0.588  & 0.205  & 0.154 \\
& 1 & 0.697  & 0.222  & 0.170 \\
& 5 & 0.731 & 0.255  & 0.201 \\
& 10 & 0.745  & 0.269  & 0.215\\
& 20 & 0.768  & 0.291 & 0.237 \\
& $20_{unbal}$ & 0.759  & 0.276  & 0.222 \\
& 40 & \textbf{0.791}  & \textbf{0.307}  & \textbf{0.251} \\
\hline
\multirow{3}*{2975\cite{tripathy2018learning}}
& 30 & 0.696 & 0.224 & 0.173\\
& 50 & 0.702 & 0.226 & 0.179\\
& 70 & 0.725 & 0.244 & 0.191\\
\hline
2975 & - & 0.569 & 0.205 & 0.156 \\
- & 2975 & 0.878 & 0.497 & 0.402\\
\hline
\end{tabular}
}
\caption{Comparison of performance of models trained for photos$\xrightarrow{}$labels task using unpaired samples with the addition of a few paired samples (Mean values from 4 runs).}
\label{tab:paired-addition}
\end{table}

\begin{figure*}
\centering
    \includegraphics[width=1.0\linewidth]{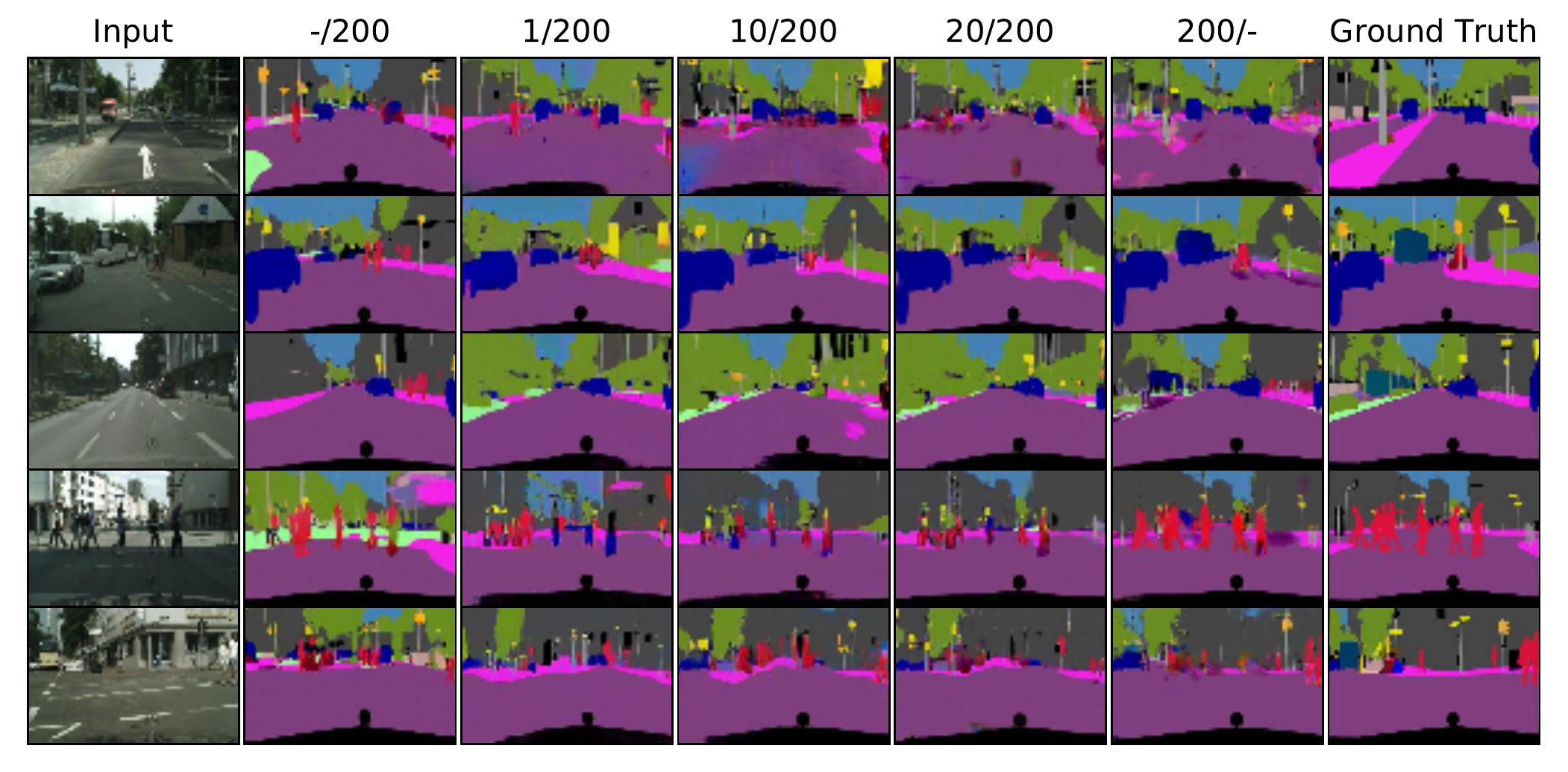}
    \caption{Qualitative performance on the photos$\xrightarrow{}$labels task for a few selected test examples. The leftmost and rightmost columns are the input images and ground truth semantic labels respectively. The 2nd column from the left and right represent CycleGAN and pix2pix models, respectively, whereas the ones in between represent our hybrid model. The numbers at the top indicate the number of paired and unpaired samples, respectively, used for training.}
\label{fig:qualitative_cityscapes}
\end{figure*}

\subsubsection{Assisting unpaired models with paired samples}

We look at the problem of using paired samples to facilitate training of unpaired image translation models. We choose 200 unpaired image samples using the k-medoids criteria and train a translation model using them. This acts as a lower baseline which we intend to improve upon by including only a small subset of paired samples. We train models by adding 1, 5, 10, 20 and 40 paired samples, respectively, to the 200 unpaired samples. For the upper baseline, we consider a model trained on all paired samples in the dataset. The results are shown in Table~\ref{tab:paired-addition}. We omit the standard deviation values, which were smaller than 0.01 for most models. We observe that adding a single paired sample improves the model significantly and makes it better than the one trained using all 2975 samples in the unpaired setting. We also observe that using unpaired samples along with a few paired ones improves the performance over the model trained using only the limited paired samples, as seen in Table~\ref{tab:results-kmedoids-vs-random-segmentation}.

\subsubsection{Balancing is critical}
It should be noted that the paired and unpaired samples are unevenly balanced with the number of unpaired samples orders of magnitude higher than the paired ones. Thus, by simply iterating over the datasets in each epoch, number of updates from unpaired samples will be much higher than the number of updates from paired samples. In order to alleviate this problem, we make copies of the paired samples so that they match in total number to the unpaired ones. This results in better performance as can be seen in Table~\ref{tab:paired-addition}, where we evaluate models trained using 20 paired samples in an unbalanced way ($20_{unbal}$). The difference in performance by not balancing was even more pronounced when we used fewer amount of paired samples for training.

\subsubsection{Selection of pairs vs. performance}
We again show that choosing paired samples using our procedure is beneficial, even for the case of assisting training with unpaired samples. For this, we train models where different single paired samples are chosen randomly ($1_{rand}$) and compare their performance with the model trained with a single paired sample selected using our k-medoids strategy. It can be seen that our selection method leads to a significantly greater improvement in performance as compared to the random selection method.

\subsubsection{Qualitative evaluation}
Fig.~\ref{fig:qualitative_cityscapes} shows a qualitative comparison of different models on the photos$\xrightarrow{}$labels task. We notice that the CycleGAN is often unable to successfully translate labels such as sky, vegetation and building. Adding a single paired sample resolves this issue in many cases, and as expected, the overall performance improves as more paired samples are considered for training the model.

\subsubsection{Using synthetic semantic labels}

\begin{figure}
\centering
    \includegraphics[width=1.0\columnwidth]{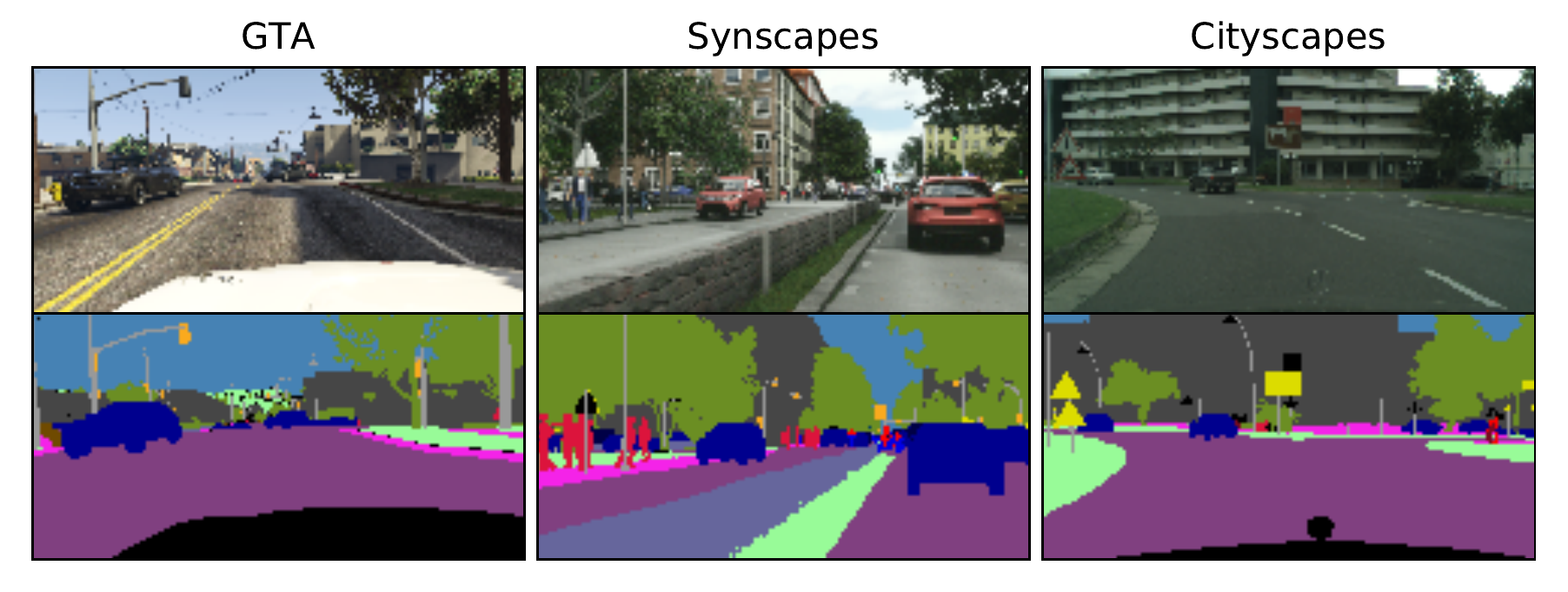}
    \caption{Top row: Scene images. Bottom row: Semantic Labels; for the GTA, Synscapes and Cityscapes datasets, respectively.}
\label{fig:datasets}
\end{figure}

There exist datasets which have synthetic semantic labels such as GTA~\cite{Richter_2016_ECCV} and Synscapes\cite{wrenninge2018synscapes}. Fig.~\ref{fig:datasets} shows the scene photographs and corresponding semantic labels from these datasets. Compared to the Cityscapes dataset, the Synscapes dataset has a similar content, structure and distribution of classes whereas for the GTA dataset they are dissimilar.

\begin{figure*}
\centering
    \includegraphics[width=1.0\linewidth]{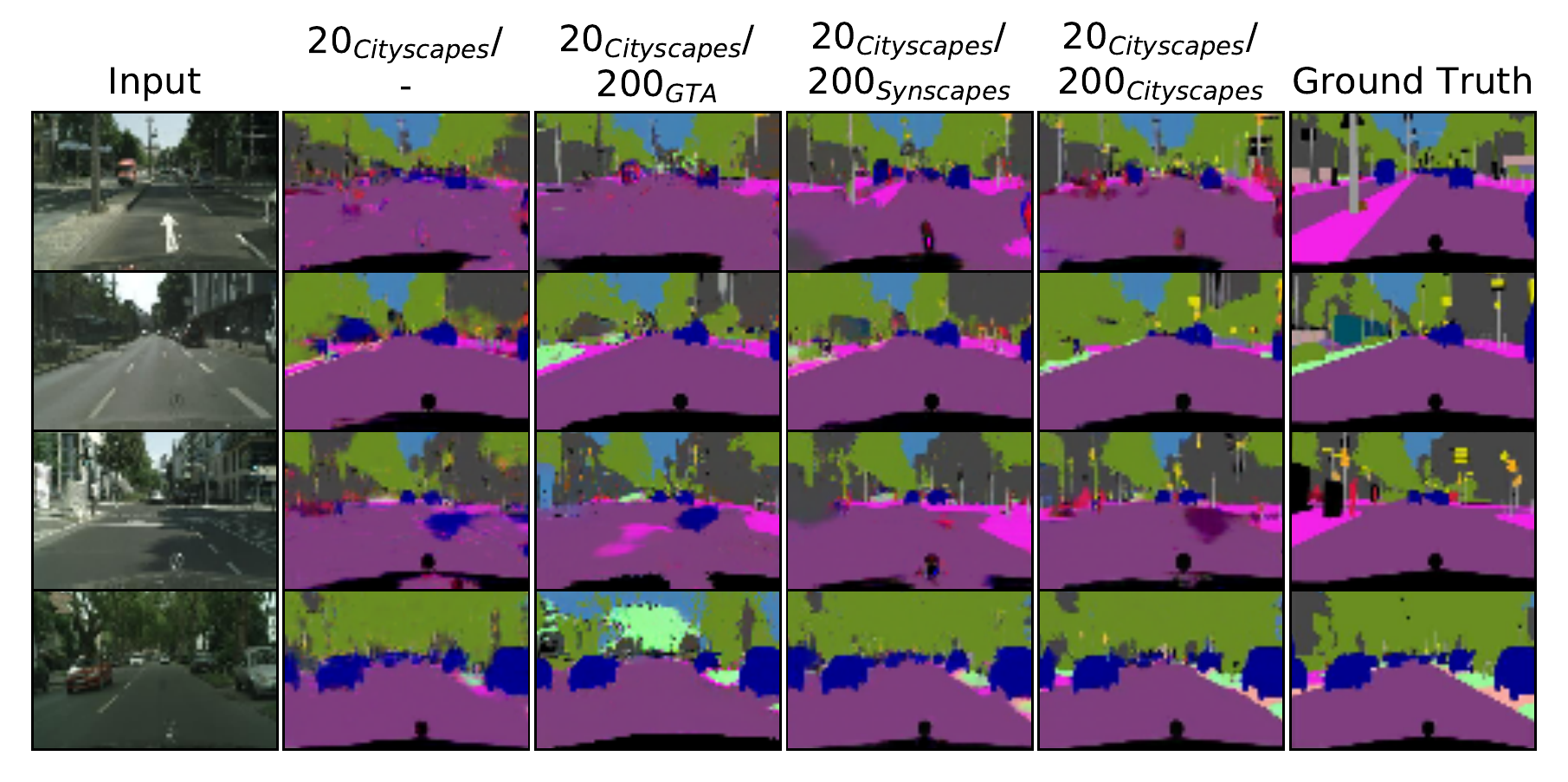}
    \caption{Qualitative performance on the photos$\xrightarrow{}$labels task for a few selected test examples. The leftmost and rightmost columns are the input images and ground truth semantic labels respectively. The 2nd column from the left represents our model trained with 20 paired samples only from Cityscapes, while the 3rd, 4th and 5th columns from the left represent models trained with a combination of 20 paired samples from Cityscapes and 200 unpaired samples with real images from Cityscapes and semantic labels from GTA, Synscapes and Cityscapes datasets, respectively.
    }
\label{fig:qualitative_cityscapes_reb}
\end{figure*}

We demonstrate the success (and shortcomings) of our approach by training two different models with 20 paired samples from Cityscapes and 200 unpaired samples.
For the unpaired samples, the real images are taken from Cityscapes and the semantic labels are taken from the GTA and Synscapes datasets, respectively, for the two models.
For reference, we also show one model trained using only 20 paired samples and another one trained using 200 unpaired samples from Cityscapes along with these 20 paired samples. They act as the lower and upper baselines, respectively.
Fig.~\ref{fig:qualitative_cityscapes_reb} shows the qualitative results and Table~\ref{tab:segmentation-synthetic} shows the corresponding quantitative results. It can be seen that using the semantic labels from the GTA dataset does not lead to significant improvement and in some cases also introduces some artifacts because of the difference in content compared to Cityscapes. However, for the Synscapes dataset, we see considerable improvement, signifying the effectiveness of our approach. The trend was similar using 1, 5, 10 and 40 paired samples but the results have been omitted due to space constraints. They can be found in the supplementary material.


\begin{table}[ht]
\setlength{\tabcolsep}{4.0pt} 
\centering
\resizebox{\linewidth}{!}
{
\begin{tabular}[\linewidth]{|c|c||c c c||}
\hline
Paired & Unpaired & Pixel Acc. & Mean Acc.& Mean IoU\\
\hline\hline
\multirow{4}*{20} & - & 0.746 & 0.242 & 0.196 \\\cline{2-5}
& $200_{GTA}$ & 0.703 & 0.259 & 0.185 \\
& $200_{Synscapes}$ & \textbf{0.761} & \textbf{0.282} & \textbf{0.224} \\\cline{2-5}
& $200_{Cityscapes}$ & 0.768 & 0.291 & 0.237 \\
\hline
\end{tabular}
}
\caption{Comparison of performance of models trained for photos$\xrightarrow{}$labels task using 20 paired samples from the Cityscapes dataset and, where applicable, 200 unpaired samples with real images from Cityscapes and semantic labels from GTA, Synscapes and Cityscapes datasets, respectively (Mean values from 4 runs).}
\label{tab:segmentation-synthetic}
\end{table}

\subsection{Satellite to Maps}

\begin{figure*}
\centering
    \includegraphics[width=1.0\linewidth]{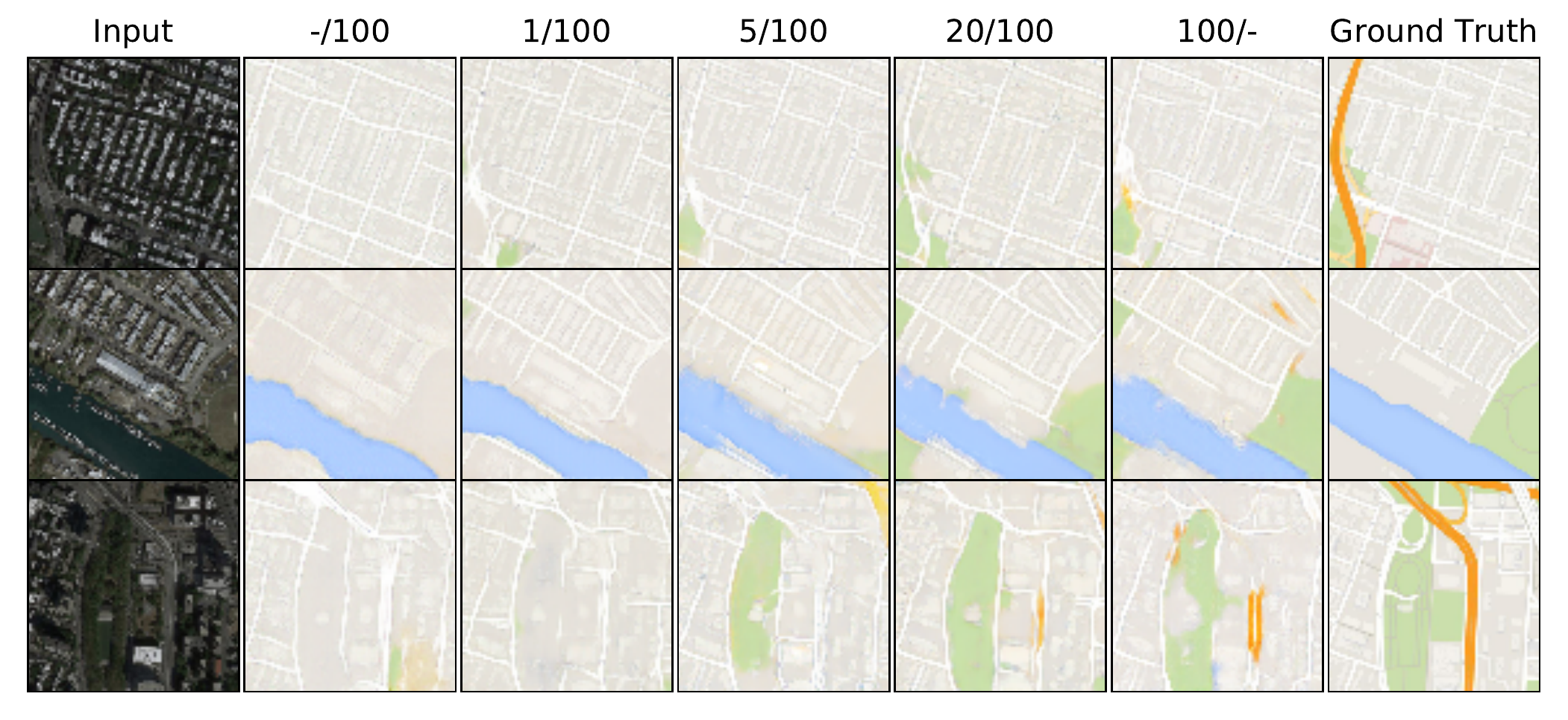}
    \caption{Qualitative performance on the satellite$\xrightarrow{}$map task for a few selected test examples. The leftmost and rightmost columns are the input images and ground truth semantic labels respectively. The 2nd column from the left and right represent CycleGAN and pix2pix models, respectively, whereas the ones in between represent our hybrid model. The numbers at the top indicate the number of paired and unpaired samples, respectively, used for training.}
\label{fig:qualitative_maps}
\end{figure*}

\subsubsection{Dataset}
The Maps dataset~\cite{Isola_2017_CVPR} consists of pairs of aerial satellite images and their corresponding maps, which is split into 1096 training pairs and 1098 pairs for testing.

We consider the task of translating from satellite images to maps. This task can also be considered as a semantic segmentation problem where the objective is to assign labels such as road, highway, building, water, vegetation etc. to each pixel. For this, we produce translated images for each sample in the test set.

\subsubsection{Evaluation protocol}
Similar to \cite{liu2017unsupervised}, we perform a quantitative evaluation by calculating the per pixel absolute difference between the translated image and the corresponding ground truth. If the maximum of this color difference across all three channels is below 20, we consider the pixel to be translated correctly. We report the average pixel accuracy across all the test samples.

\subsubsection{Settings and results}
For training models, we first select 100 samples using our strategy and establish baselines by assessing the models trained in completely paired and unpaired fashions. We also train using the entire training set in order to establish higher baselines. Table~\ref{tab:maps-evaluation} shows the performance of our model using different number of paired and unpaired samples on this task. We notice an increase in accuracy when we incorporate our training process compared to when using only unpaired samples. Surprisingly, all our hybrid models utilizing 100 unpaired samples and a small subset of paired samples perform comparable with or better than a purely supervised model which uses all 100 paired samples. A possible explanation is that the supervised model may overfit to the training samples, whereas the loose constraint while using unpaired samples may lead to better generalization.

\begin{table}[htp]
\begin{center}
\begin{tabular}{|c | c || c ||}
\hline
Unpaired & Paired & Pixel Acc.\\
\hline\hline
- & 100 & 0.377 $\pm$ 0.018\\
100 & - & 0.356 $\pm$ 0.016\\
\hline
\multirow{4}*{100}
& 1 & 0.405 $\pm$ 0.152 \\
& 5 & 0.379 $\pm$ 0.013\\
& 10 & 0.415 $\pm$ 0.006 \\
& 20 &  \textbf{0.431 $\pm$ 0.009} \\
\hline
1096 & - & 0.388 \\
- & 1096 & 0.439 \\
\hline
\end{tabular}
\end{center}
\caption{Comparison of performance of models trained on the satellite$\xrightarrow{}$map task using unpaired samples with the addition of a few paired samples (Mean values and standard deviations from 4 runs). }
\label{tab:maps-evaluation}
\end{table}

\subsubsection{Top performance with extremely weak supervision}
Again using as few as one paired sample in addition to one hundred unpaired samples leads to large improvements over using one hundred unpaired alone and comparable with using all the unpaired samples ($\equiv$ CycleGAN). If the number of selected paired samples is increased up to 20 in addition to one hundred unpaired samples, then, a performance comparable with using all the paired samples (1096) in full supervision ($\equiv$ pix2pix) is achieved.

\subsubsection{Qualitative evaluation}
Fig.~\ref{fig:qualitative_maps} shows visual results for the satellite to maps translation problem. It can be seen that adding paired samples leads to overall improvement in visual quality of the translated images, although it is still prone to errors.

\section{Conclusion}
\label{sec:discussion}

We presented a framework
which can use both paired and unpaired samples for training image-to-image translation models.
We also proposed a method to select samples for which pairs are to be acquired given a fixed budget, which performs significantly better than random selection. Our experimental results demonstrate how using just one judicially selected paired sample leads to significant gains in performance. Our extremely weak supervised image-to-image translation solution validated on semantic segmentation tasks represents a strong step towards bridging the gap in performance between unsupervised and supervised image-to-image paradigms with huge practical impact.

\paragraph{Acknowledgments. }
This work was partly supported by ETH General Fund and by Nvidia through a GPU grant.

{\small
\bibliographystyle{./ieee}
\bibliography{./main}
}


\section{Supplementary Material}

\subsection{Qualitative Results : Cityscapes Photos$\xrightarrow{}$Labels}

\begin{figure*}[htp]
\centering
    \includegraphics[width=1.0\linewidth]{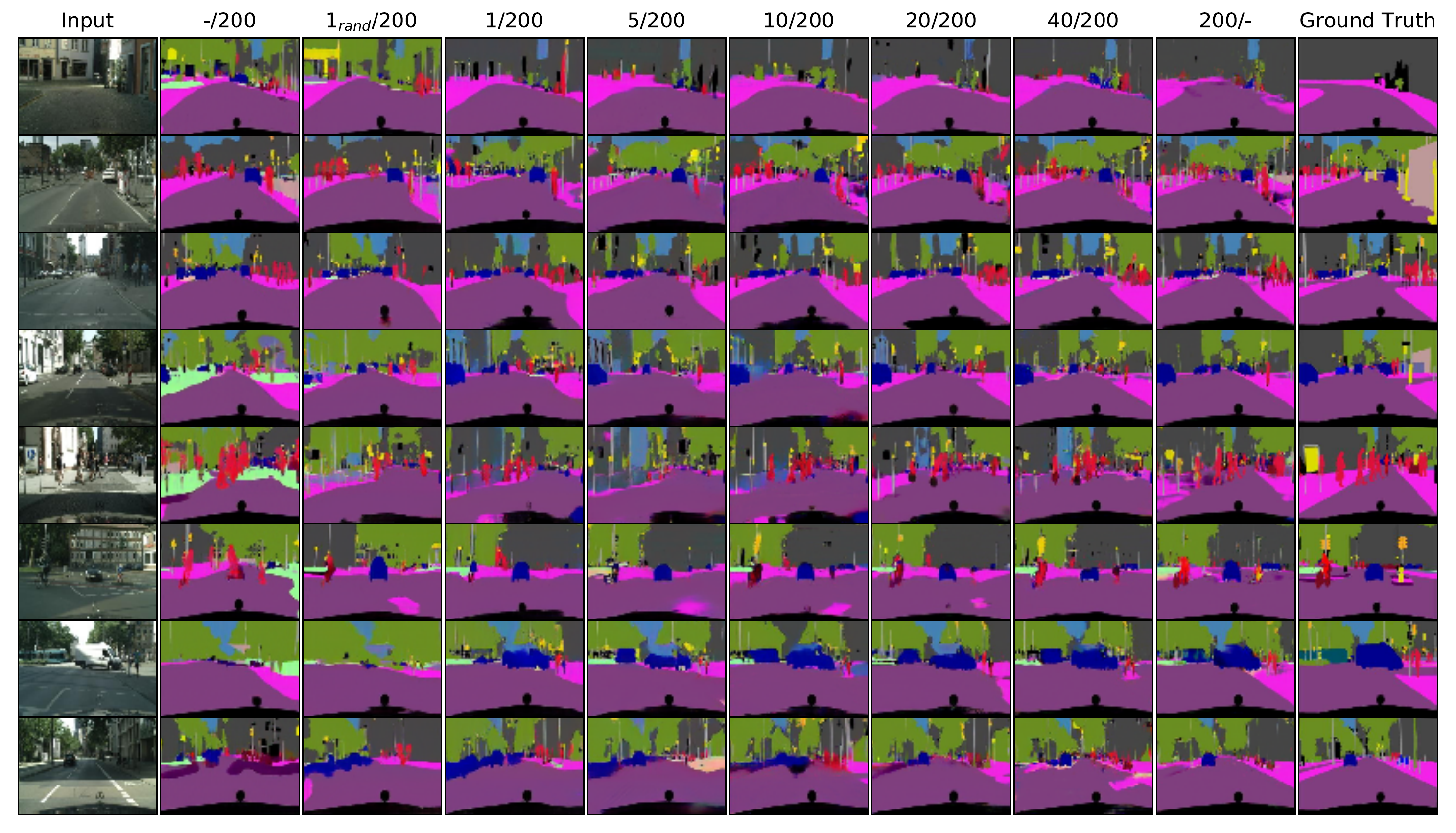}
    \caption{Qualitative performance on the photos$\xrightarrow{}$labels task for a few selected test examples. The leftmost and rightmost columns are the input images and ground truth images respectively. The 2nd column from the left and right represent CycleGAN and pix2pix models, respectively, whereas the ones in between represent our hybrid model. The numbers at the top indicate the number of paired and unpaired samples, respectively, used for training. All paired samples are chosen using our k-medoids strategy for training apart from the 3rd column where 1 sample is chosen randomly.}
\label{fig:qualitative_cityscapes_sup}
\end{figure*}

Fig.~\ref{fig:qualitative_cityscapes_sup} shows additional qualitative comparison of different models on the photos$\xrightarrow{}$labels task.

\subsection{Quantitative Evaluation : Cityscapes Photos$\xrightarrow{}$Labels}

\subsubsection{Ablation Study}

We compare the performance of our model without the cycle-loss and identity-loss respectively on the cityscapes photos$\xrightarrow{}$labels task, using 20 paired samples and 200 unpaired ones. Table \ref{tab:ablation-study} shows that both the cycle loss and identity loss are important for our task.

\begin{table}[ht]
\centering
{
\begin{tabular}[\linewidth]{|c||c c c||}
\hline
Model  & Pixel Acc. & Mean Acc. & Mean IoU\\
\hline\hline
Full & \textbf{0.768} & \textbf{0.291} & \textbf{0.237}\\
w/o $cycle$ & 0.767 & 0.279 & 0.227 \\
w/o $idt$ & 0.759 & 0.277 & 0.224 \\
\hline
\end{tabular}
}
\caption{Average performance across 4 runs of different models trained for photos$\xrightarrow{}$labels task using 20 paired and 200 unpaired samples from the cityscapes dataset.}
\label{tab:ablation-study}
\end{table}

\subsubsection{Comparison of our I2I model vs completely supervised semantic segmentation methods}

We compare our semantic segmentation model with FCN-8s\cite{long2015fully} using a publicly available implementation\footnote{https://github.com/zijundeng/pytorch-semantic-segmentation}. We use the default hyperparameters and train for the same amount of time as our method ($\sim$5 hours).  It should be noted that the FCN-8s segmentation model only uses the paired samples whereas ours uses both paired and unpaired samples. Table \ref{tab:fcn-vs-ours} shows the evaluation results. It can be seen that we achieve better performance.

\begin{table}[htp]
\setlength{\tabcolsep}{2pt}
\centering
\resizebox{\linewidth}{!}
{
\begin{tabular}{|c | c | c || c c c ||}
\hline
Method & Paired & Unpaired & Pixel Acc. & Mean Acc. & Mean IoU\\
\hline\hline
FCN-8s\cite{long2015fully} & 20 & - & 0.544 & 0.158 & 0.083\\
Ours & 20 & 200 & \textbf{0.768} & \textbf{0.291} & \textbf{0.237}\\
\hline
\end{tabular}
}
\caption{Comparison of performance of models trained using FCN-8s (20 paired samples) and our method (20 paired samples and 200 paired samples) for the photos$\xrightarrow{}$labels task.}
\label{tab:fcn-vs-ours}
\end{table}

\subsection{Synthetic labels in unpaired training setting}

\begin{figure*}
\centering
    \includegraphics[]{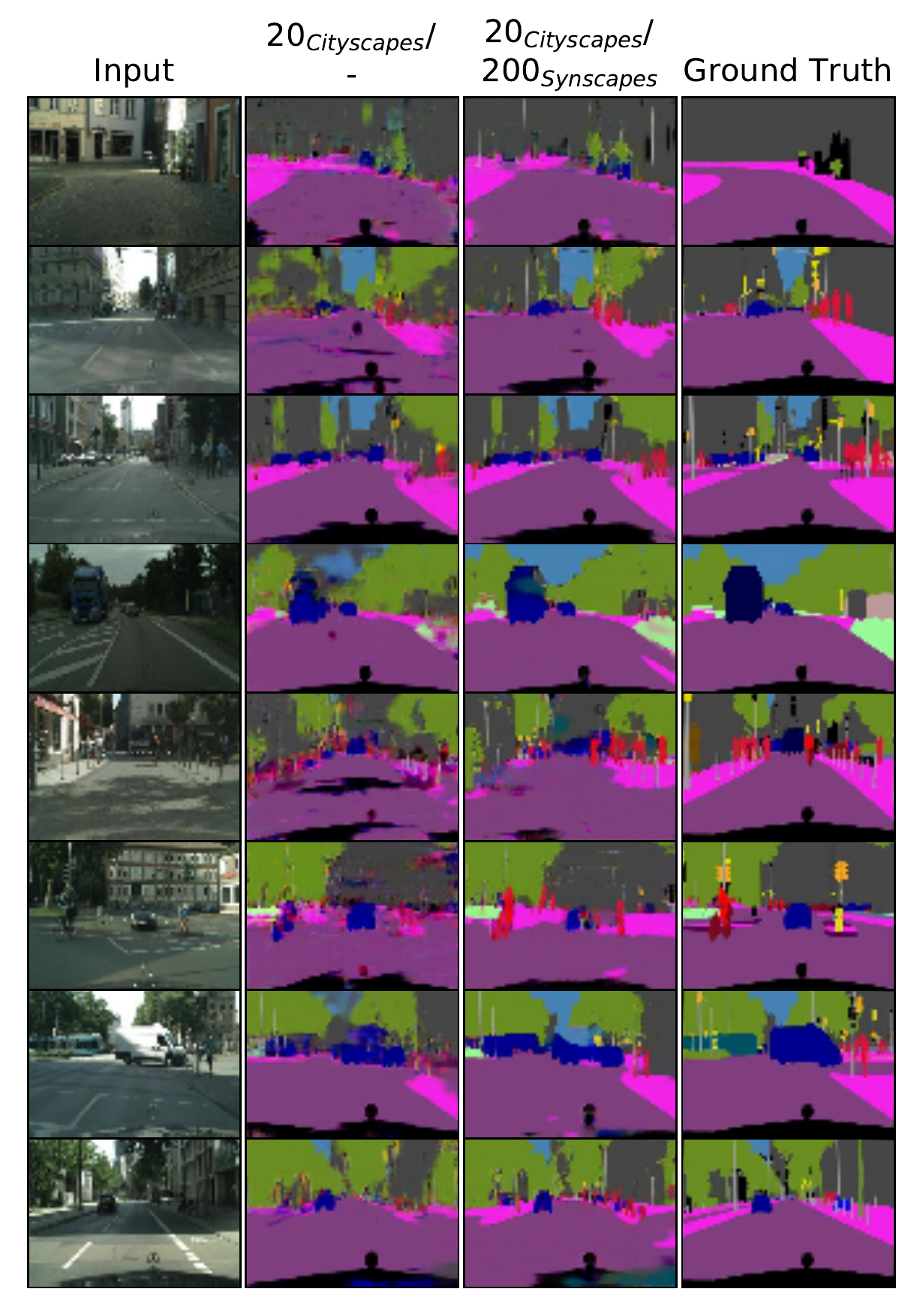}
    \caption{Qualitative performance on the photos$\xrightarrow{}$labels task for a few selected test examples. The leftmost and rightmost columns are the input images and ground truth semantic labels respectively. The 2nd column from the left represents pix2pix model trained with 20 paired samples from Cityscapes, while the 3rd column from the left represents our hybrid model trained with a combination of 20 paired samples from Cityscapes and 200 unpaired samples with real images from Cityscapes and semantic labels from Synscapes dataset.
    }
\label{fig:qualitative_cityscapes_syn_ext}
\end{figure*}

We present more results for models where we use synthetic semantic annotations in the unpaired setting (see Sec 4.1.8 in paper). Table \ref{tab:segmentation-synthetic-ext} shows quantitative results and Fig. \ref{fig:qualitative_cityscapes_syn_ext} shows qualitative results. It can be seen that we can achieve consistent performance gains by utilizing the semantic labels from the Synscapes~\cite{wrenninge2018synscapes} dataset in the unpaired setting over models which are trained in the purely paired setting.

\begin{table}[ht]
\setlength{\tabcolsep}{4.0pt} 
\centering
\resizebox{\linewidth}{!}
{
\begin{tabular}[\linewidth]{|c|c||c c c||}
\hline
Paired & Unpaired & Pixel Acc. & Mean Acc.& Mean IoU\\
\hline\hline
\multirow{4}*{1} & - & 0.567 & 0.160 & 0.114 \\
& $200_{GTA}$ & 0.614 & 0.215 & 0.137 \\
& $200_{Synscapes}$ & 0.674 & \textbf{0.249} & \textbf{0.179} \\
& $200_{Cityscapes}$ & \textbf{0.697}  & 0.222  & 0.170 \\
\hline
\multirow{4}*{5} & - & 0.636 & 0.189 & 0.143 \\
& $200_{GTA}$ & 0.637 & 0.217 & 0.144 \\
& $200_{Synscapes}$ & 0.697 & 0.253 & 0.188 \\
& $200_{Cityscapes}$ & \textbf{0.731} & \textbf{0.255} & \textbf{0.201} \\
\hline
\multirow{4}*{10} & - & 0.720 & 0.229 & 0.174 \\
& $200_{GTA}$ & 0.646 & 0.228 & 0.153 \\
& $200_{Synscapes}$ & 0.737 & \textbf{0.270} & 0.209 \\
& $200_{Cityscapes}$ & \textbf{0.745}  & 0.269 & \textbf{0.215} \\
\hline
\multirow{4}*{20} & - & 0.746 & 0.242 & 0.196 \\
& $200_{GTA}$ & 0.703 & 0.259 & 0.185 \\
& $200_{Synscapes}$ & 0.761 & 0.282 & 0.224 \\
& $200_{Cityscapes}$ & \textbf{0.768} & \textbf{0.291} & \textbf{0.237} \\
\hline
\multirow{4}*{40} & - & 0.781 & 0.278 & 0.236 \\
& $200_{GTA}$ & 0.745 & 0.291 & 0.216 \\
& $200_{Synscapes}$ & 0.786 & \textbf{0.309} & 0.248 \\
& $200_{Cityscapes}$ & \textbf{0.791}  & 0.307  & \textbf{0.251}\\
\hline
\end{tabular}
}
\caption{Comparison of performance of models trained for photos$\xrightarrow{}$labels task using different number of paired samples from the Cityscapes dataset and, where applicable, 200 unpaired samples with real images from Cityscapes and semantic labels from GTA, Synscapes and Cityscapes datasets respectively (Mean values from 4 runs).}
\label{tab:segmentation-synthetic-ext}
\end{table}

\subsection{Quantitative Evaluation : Cityscapes Labels$\xrightarrow{}$Photos}

For the labels$\xrightarrow{}$photos task, the assumption is that a high quality generated image would lead to the original image when passed through a well-trained semantic segmentation model.
We use the pretrained FCN~\cite{long2015fully} semantic segmentation model from the pix2pix~\cite{Isola_2017_CVPR} code.
We resize the generated fake images to the original resolution of Cityscapes ($1024\times2048$px).
We then apply the pretrained model to obtain the semantic label and compare it with the ground truth image from cityscapes to compute the metrics.

\subsubsection{Comparison of paired I2I translation using different subsets}

Table \ref{tab:results-kmedoids-vs-random} shows performance of models trained using subsets of different number of samples selected using k-medoids and random criteria (average of 10 runs) on the labels$\xrightarrow{}$photos task. It can be seen that for this task as well, models trained with samples chosen using the k-medoids criteria outperform the ones trained with samples selected randomly.

\begin{table}[ht]
\setlength{\tabcolsep}{4pt} 
\centering
\resizebox{\linewidth}{!}
{
\begin{tabular}[\linewidth]{|c|c||c c c||}
\hline
Samples & Selection & Pixel Acc. & Mean Acc. & Mean IoU\\
\hline\hline
\multirow{2}*{1} & random & 0.494 & 0.156 & 0.099 \\
& k-medoids & \textbf{0.588} & \textbf{0.187} & \textbf{0.125} \\
\hline
\multirow{2}*{5} & random & 0.612 & 0.203 & 0.144 \\
& k-medoids & \textbf{0.645} & \textbf{0.202} & \textbf{0.147} \\
\hline
\multirow{2}*{10} & random & 0.695 & 0.228 & 0.171 \\
& k-medoids & \textbf{0.723} & \textbf{0.240} & \textbf{0.186} \\
\hline
\multirow{2}*{20} & random & 0.714 & 0.232 & 0.179 \\
& k-medoids & \textbf{0.730} & \textbf{0.241} & \textbf{0.187} \\
\hline
\multirow{2}*{40} & random & 0.718 & 0.229 & 0.178 \\
& k-medoids & \textbf{0.737} & \textbf{0.232} & \textbf{0.183} \\
\hline\hline
2975 & - & 0.760 & 0.253 & 0.200\\
\hline
\end{tabular}
}
\caption{Average performance of models trained for labels$\xrightarrow{}$photos task using a subset of different sizes from the Cityscapes dataset (random vs k-medoids selection). For k-medoids selection, we take an average across 10 runs using the same samples, whereas for random selection, we take average for 10 runs using different random samples each time}
\label{tab:results-kmedoids-vs-random}
\end{table}

\subsubsection{Comparison of unpaired models, assisted by a few different number of paired samples}

We train models for the labels$\xrightarrow{}$image translation task using 200 unpaired samples, and try to assist the models by a few paired samples. The corresponding results are shown in Table \ref{tab:paired-addition-inv}. We notice a significant increase in performance by only adding a single sample which is already comparable to the model trained with the complete paired dataset.

\begin{table}[htp]
\setlength{\tabcolsep}{2pt}
\centering
\resizebox{\linewidth}{!}
{
\begin{tabular}{|c | c || c c c||}
\hline
& & & label$\xrightarrow{}$photo &  \\
Unpaired & Paired & Pixel Acc. & Mean Acc. & Mean IoU\\
\hline\hline
- & 200 & 0.762  & 0.254  & 0.203 \\
200 & - & 0.618  & 0.191  & 0.148 \\
\hline
\multirow{7}*{200 (\textbf{ours})}
& $1_{rand}$ & 0.619 & 0.215 & 0.158 \\
& 1 & 0.734 & 0.245 & 0.186 \\
& 5 & 0.741 & 0.258 & 0.197 \\
& 10 & 0.757 & \textbf{0.266} & 0.204 \\
& 20 & \textbf{0.765} & 0.264 & \textbf{0.207}\\
& $20_{unbal}$ & 0.752 & 0.257 & 0.198\\
& 40 & 0.761  & 0.260 & 0.202\\
\hline
\multirow{3}*{2975\cite{tripathy2018learning}}
& 30 & 0.66 & 0.19 & 0.14 \\
& 50 & 0.65 & 0.19 & 0.13 \\
& 70 & 0.67 & 0.19 & 0.14 \\
\hline
2975 & - & 0.583 & 0.187 & 0.138 \\
- & 2975 & 0.760 & 0.253 & 0.200 \\
\hline
\end{tabular}
}
\caption{Comparison of performance of models trained for labels$\xrightarrow{}$photos task using unpaired samples with the addition of a few paired samples (Mean values from 4 runs).}
\label{tab:paired-addition-inv}
\end{table}

\subsection{Qualitative Results : Satellite$\xrightarrow{}$Map}

\begin{figure*}[htp]
\centering
    \includegraphics[width=1.0\linewidth]{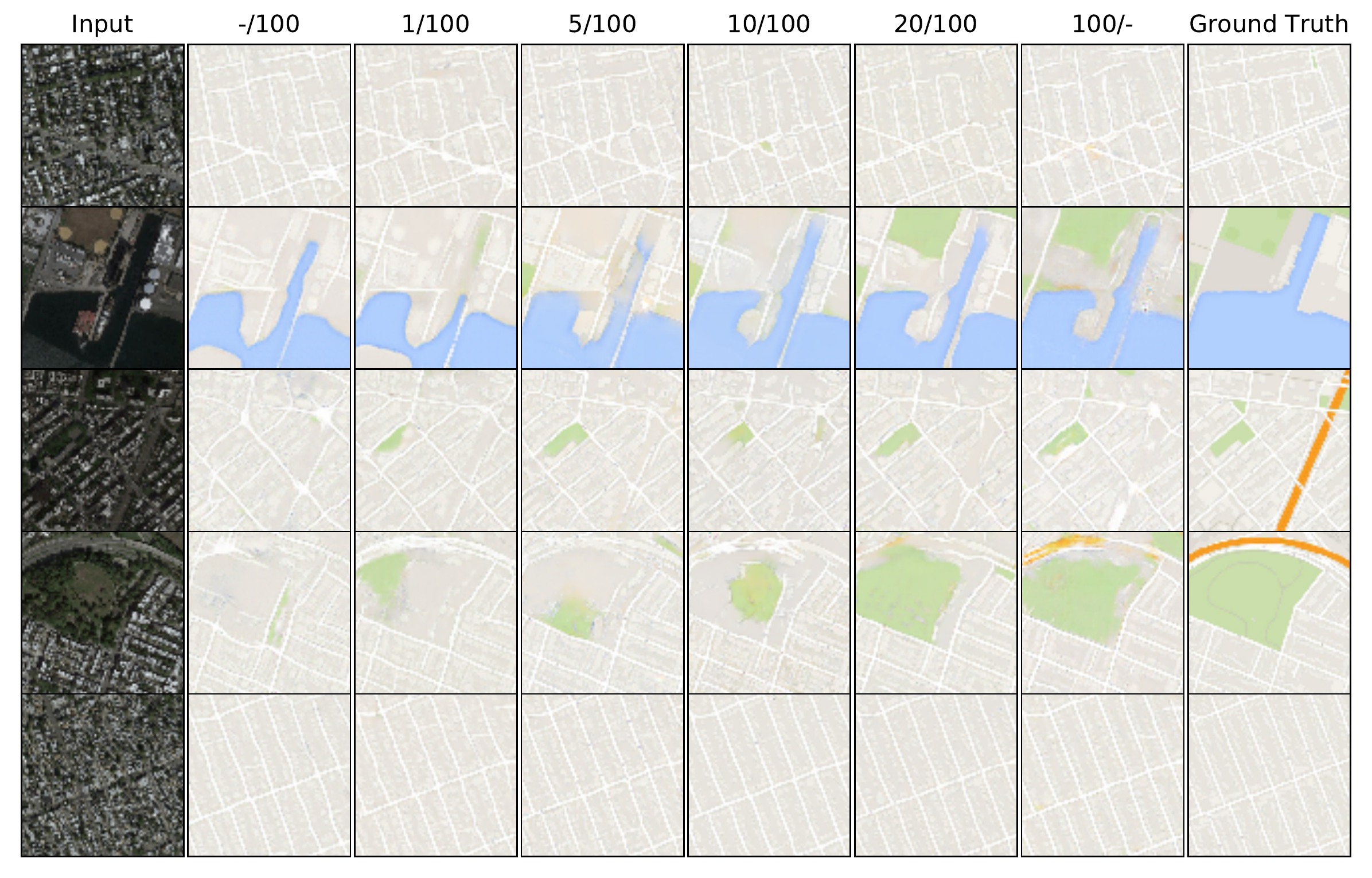}
    \caption{Qualitative performance on the satellite$\xrightarrow{}$map task for a few selected test examples. The leftmost and rightmost columns are the input images and ground truth images respectively. The 2nd column from the left and right represent CycleGAN and pix2pix models, respectively, whereas the ones in between represent our hybrid model. The numbers at the top indicate the number of paired and unpaired samples, respectively, used for training.}
\label{fig:qualitative_maps_sup}
\end{figure*}

Fig.~\ref{fig:qualitative_maps_sup} shows additional qualitative comparison of different models on the satellite$\xrightarrow{}$map task.

\end{document}